\documentclass{bmvc2k}

\title{CLAIR: CLIP-Aided Weakly Supervised Zero-Shot Cross-Domain Image Retrieval}

\addauthor{Chor Boon Tan}{tan.chor.boon@u.nus.edu}{1}
\addauthor{Conghui Hu}{conghui@nus.edu.sg}{1}
\addauthor{Gim Hee Lee}{gimhee.lee@nus.edu.sg}{1}

\addinstitution{
 School of Computing\\
 National University of Singapore\\
}

\runninghead{CB Tan, C Hu, GH Lee}{CLAIR}

\def\eg{\emph{e.g}\bmvaOneDot}
\def\ie{\emph{i.e}\bmvaOneDot}

\usepackage{xspace}
\usepackage{enumitem}
\usepackage{amssymb}
\usepackage{booktabs}
\usepackage{multirow}
\usepackage{cleveref}
\usepackage{nomencl}

\begin{document}

\maketitle

\begin{abstract}

The recent growth of large foundation models that can easily generate pseudo-labels for huge quantity of unlabeled data makes unsupervised Zero-Shot Cross-Domain Image Retrieval (UZS-CDIR) less relevant. In this paper, we therefore turn our attention to weakly supervised ZS-CDIR (WSZS-CDIR) with noisy pseudo labels generated by large foundation models such as CLIP. To this end, we propose CLAIR to refine the noisy pseudo-labels with a confidence score from the similarity between the CLIP text and image features. Furthermore, we design inter-instance and inter-cluster contrastive losses to encode images into a class-aware latent space, and an inter-domain contrastive loss to alleviate domain discrepancies. We also learn a novel cross-domain mapping function in closed-form, using only CLIP text embeddings to project image features from one domain to another, thereby further aligning the image features for retrieval. Finally, we enhance the zero-shot generalization ability of our CLAIR to handle novel categories by introducing an extra set of learnable prompts. Extensive experiments are carried out using TUBerlin, Sketchy, Quickdraw, and DomainNet zero-shot datasets, where our CLAIR consistently shows superior performance compared to existing state-of-the-art methods. We release our code at this \href{https://github.com/aniruok9/CLAIR}{https URL}.

\end{abstract}

\section{Introduction}
\label{sec:intro}

\begin{figure}
    \centering
    \includegraphics[width=0.9\columnwidth]{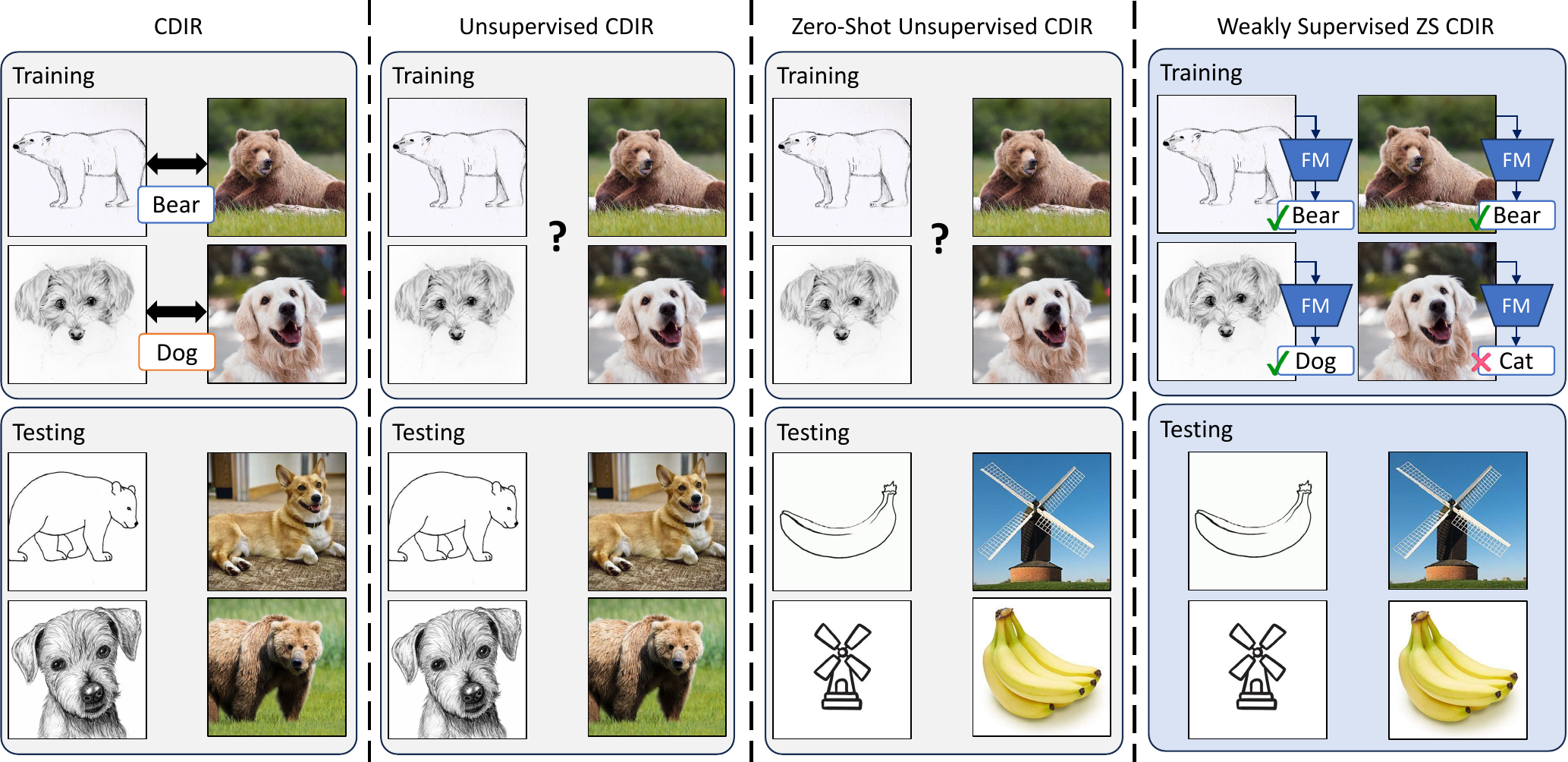}
     
     \vspace{-3mm}
     \caption{From left to right, \textbf{CDIR} assumes both class labels and paired images across domains are provided during training. In \textbf{UCDIR}, class labels and paired annotations are no longer available. \textbf{ZS-UCDIR} introduces new unseen classes during the test phase. The use of foundation model converts the problem into \textbf{WSZS-CDIR}, where the model is trained under weak supervision of noisy pseudo labels.}
     \vspace{-5mm}
  \label{fig:problem_setting}
  
\end{figure}

Cross-Domain Image Retrieval (CDIR) is the task of retrieving images from a target domain that are pertinent to a query image from a source domain. The vanilla CDIR setting relies on supervised training with category labels and/or pairing annotations from both domains, limiting its real-world applicability due to high annotation costs. \textbf{Zero-shot CDIR} \cite{Lin2023, Sain2023} addresses this by training on a closed set of categories and generalizing to unseen ones at test time. However, it still necessitates an annotated training set. \textbf{Unsupervised CDIR} \cite{Hu2022, Wang2023} aims for label-free domain alignment, but its application is limited to a closed set of seen categories, and is increasingly becoming obsolete with the availability of large foundation models such as CLIP \cite{Radford2021a} that are capable of generating pseudo labels on large amounts of unlabeled data.

In view of these limitations, we investigate the problem of Weakly Supervised Zero-Shot CDIR (WSZS-CDIR). As illustrated in Figure \ref{fig:problem_setting}, this task is a variant of the recently proposed Unsupervised Zero-Shot CDIR (UZS-CDIR) problem setting of image sketches \cite{yin2024asymmetric} (also known as Zero-Shot Sketch-Based Image Retrieval, \ie ZS-SBIR) with two important differences: 1) We aim to generalize our model across any domain instead of limiting to only sketch image domain. 2) The existence of many large foundation models has significantly reduced the difficulty in obtaining pseudo-labels for training.  Consequently, we have converted UZS-CDIR into a WSZS-CDIR problem. Nonetheless, 
several challenges remain: 1) The pseudo-labels generated from large foundation models tend to be extremely noisy. A naive use of these noisy pseudo-labels to train the model cannot yield good performance. 2) A robust method to extract semantic-aware and domain-invariant features for effective cross-domain alignment is still needed. 3) A simple training on the pseudo-labels of the available data limits the model to the class categories in the training data and does not give the model zero-shot ability to generalize on unseen categories. 

We propose CLAIR in this paper, which stands for \underline{\textbf{CL}}IP-\underline{\textbf{A}}ided Weakly Supervised framework for Zero-Shot Cross-Domain \underline{\textbf{I}}mage \underline{\textbf{R}}etrival. Our CLAIR leverages CLIP to generate noisy pseudo-labels of the unlabeled training data from any two domains. We mitigate the noisy pseudo-labels issue by computing the confidence scores of the pseudo-labels assigned to the training images in both query and target domains based on the similarity between their CLIP text and image features. To effectively align distinct image domains into a shared class-aware latent space, we design inter-instance and inter-cluster contrastive losses for encoding semantic features, and an inter-domain contrastive loss to alleviate domain discrepancies. We further improve domain alignment by deriving a closed-form mapping function using easily generated paired cross-domain CLIP text prompts and directly applying it to image features. Lastly, we enhance zero-shot generalization ability by concatenating a set of learnable prompts \cite{Wang2022} with image features into the encoder for training.

Our CLAIR framework undergoes end-to-end training that alternates between optimizing the parameters of the domain alignment encoder network and the pseudo-labels in a refinement process. During inference, we pass images of both domains through the domain alignment encoder to obtain image features for ranking.

Our main contributions can be summarized as follows:
\begin{enumerate}[noitemsep]
    \item We introduce the problem setting of Weakly Supervised Zero-Shot Cross-Domain Image Retrieval (WSZS-CDIR), and develop a novel framework CLAIR that tackles all the challenges of this setting.
    \item We propose a novel domain mapping function that requires no training images to reduce the domain gap.
    \item We propose a multi-granular contrastive learning loss that minimizes feature distance between semantically similar samples across instances, clusters and domains.
    \item We validate the efficacy of our proposed framework with extensive experiments on the TUBerlin Extended, Sketchy Extended, Quickdraw Extended and DomainNet Zero-Shot datasets. The latter is a modified train/test split of DomainNet, specifically tailored by us for the evaluation of zero-shot tasks spanning multiple domains.
\end{enumerate}
\section{Related Work}
\label{sec:related}

\paragraph{Unsupervised Cross-Domain Image Retrieval.}
\cite{Hu2022a} uses optimal transport to establish paired domain correspondence. \cite{Hu2022} combines cluster-wise contrastive loss and a cross-domain distance of distance loss to reduce the domain gap, while \cite{Wang2023} minimizes cross-domain discrepancy of both classifiers by enforcing consistent predictions of the same image. However, UCDIR approaches do not work under the zero-shot setting.

\vspace{-3mm} \paragraph{Zero-shot Sketch-based Image Retrieval.}
While early works relied on word embeddings \cite{Dey2019, zhang2020zero}, later approaches have moved away from such semantics. \cite{Sain2022, Hudson2023} adopts test-time training to improve domain alignment before inference. \cite{Tian2022a, Lin2023} applied Vision Transformers (ViTs) \cite{Dosovitskiy2021} to train a Zero-shot model.  \cite{Sain2023} is the first to utilize CLIP for ZS-SBIR tasks. However, these approaches all require a training dataset with explicit categorical labels or paired images across domains. More recently, \cite{yin2024asymmetric} proposes self-distillation with DINO \cite{oquab2024dinov2learningrobustvisual} to tackle ZS-SBIR with an Unsupervised constraint (UZS-SBIR).

\vspace{-3mm} \paragraph{Contrastive Language Pre-training.}
CLIP \cite{Radford2021a} is a cross-modal visual-language foundation model trained on a substantial dataset comprising approximately 400 million paired samples. Its robustness and generalizability have been showcased in various downstream applications such as Image Classification \cite{Abdelfattah2023}, ZS-SBIR \cite{Sain2023}, Zero-shot Semantic Segmentation \cite{Zhou2022}, prompt-guided semantic editing \cite{Abdal2022}, \textit{etc}. We adopt CLIP to generate pseudo labels for weak supervision and for zero-shot generalization across domains.

\vspace{-3mm} \paragraph{Prompt Learning.}
Fundamentally, the concept involves modifying inputs to add context to foundation models, improving performance. \cite{Wang2022} optimized a frozen model with learnable prompts for specific tasks. \cite{Jia2022} showed that adding a small amount of learnable prompts boosts ViT performance across various downstream tasks. \cite{Singha2023, Ge2022} demonstrated the efficiency of prompt-learning CLIP for Unsupervised Domain Adaptation (UDA). We integrate prompt learning to improve our model's ability to retrieve unseen classes. 

\section{Problem Definition}
\label{sec:method}

We follow the settings of UZS-CDIR. Given a query image $I_{i}^A\in\mathbb{R}^{h\times w\times c}$ from domain A, our task is to retrieve images of the same category in domain B, where $i$ is the index of the image. Different from the traditional zero-shot setting where training images from both domains are labeled, our setting utilizes only the unlabeled training images $\mathcal{I}^A_{train}=\{I^A_{i} \}^N_{i=1}$ and $\mathcal{I}^B_{train}=\{I^B_{j} \}^M_{j=1}$, and the set of training categories names $\gamma_{train}$. As defined in \cite{miyai2024generalized} for CLIP-based zero-shot problems, we evaluate our model on novel categories, \ie, $\gamma_{train} \cap \gamma_{test} = \varnothing$. Additionally, training images across the two domains are unpaired. We convert UZS-CDIR into WSZS-CDIR by generating noisy pseudo labels of the unlabeled training data using CLIP.

\begin{figure}[t]
  \centering
  \includegraphics[width=0.95\columnwidth]{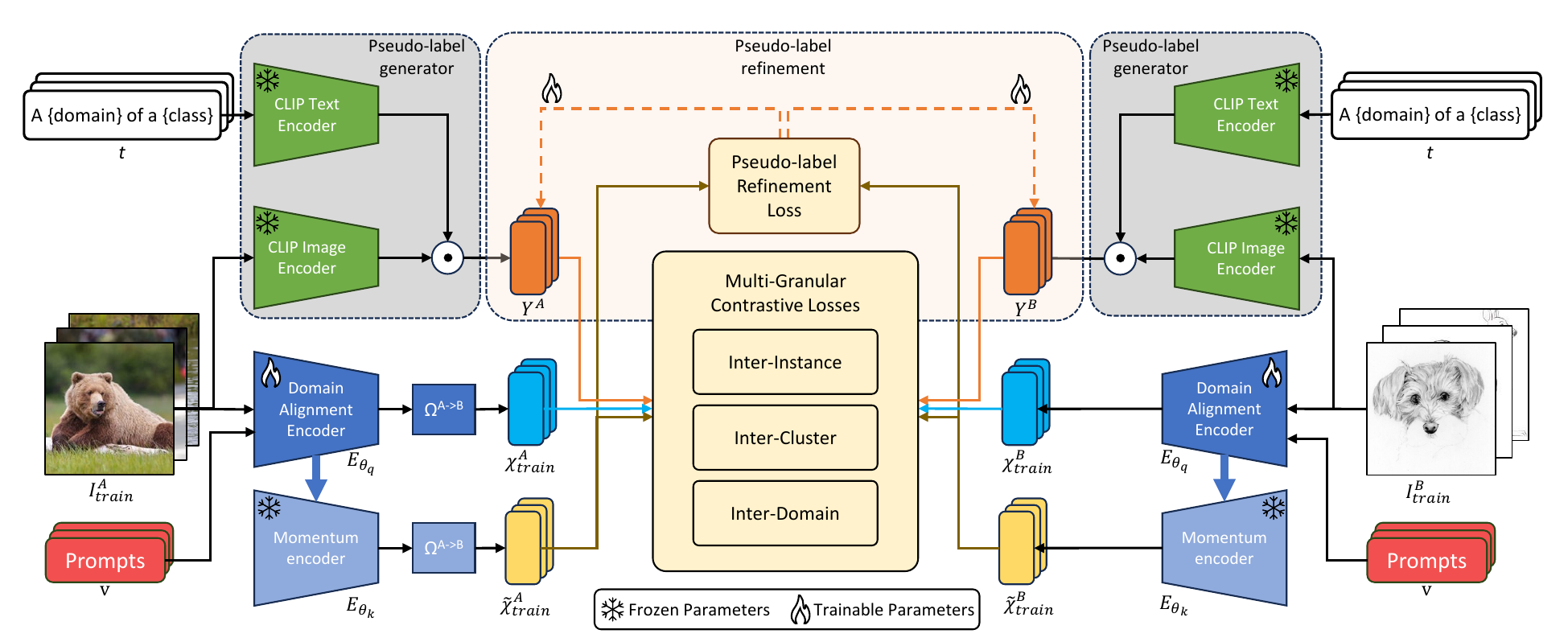}
  \vspace{-0.5mm}
  \caption{An overview of our proposed CLAIR. 
  \textbf{Before training}, CLIP is used to initialize pseudo-labels. A Cross-domain mapping function $\Omega^{A\rightarrow B}$ provides initial feature alignment, with parameters updated during training. Learnable prompts $\text{v}$ are input along with images to the domain alignment encoder $E_{\theta_q}$, mirrored with a momentum encoder $E_{\theta_k}$ for training stability. 
  \textbf{During training}, $E_{\theta_q}$ is optimized using contrastive losses on multiple granualities, guided by pseudo-labels and samples from the features $\chi_{train}$ generated by $E_{\theta_k}$. Pseudo-labels are updated at every epoch end via a refinement loss supervised by $\chi_{train}$, with gradient scaling biased to unconfident labels. The prompts $\text{v}$ and both $E_{\theta_q}$ and $E_{\theta_k}$ are shared across domains. 
  \textbf{During inference}, $E_{\theta_k}$ is used to produce image features. The framework is best viewed in color.  }
  
  \vspace{-5mm}
  \label{fig:model_arch}
\end{figure}

\section{Our Methodology}
Fig.~\ref{fig:model_arch} shows an illustration of our CLAIR for WSZS-CDIR. Formally, we train a domain alignment encoder $E_{\theta_q} : I_i \mapsto x_i\in\mathbb{R}^d$ that projects images from both domains to a common feature space. We denote $y_i^A \in \mathbb{R}^{K}$ as the pseudo-label vector of the embedding vector $x_i^A$ and its corresponding image $I_i^A$, where $K$ is the number of classes in the training dataset. $E_{\theta_q}$ is mirrored by a momentum encoder $E_{\theta_k} : I_i \mapsto \tilde{x}_i\in\mathbb{R}^d$, whose weights are updated at the end of every epoch by $\theta_k \leftarrow m\theta_k + (1 - m)\theta_q$, where $m$ is a hyperparameter. It provides a stabler set of embeddings to be used as loss supervision \cite{Chen2020}. During inference, image features are encoded using only $E_{\theta_k}$. 

\subsection{Pseudo-Label Initialization}
We employ CLIP text encoder $E_{t}$ to generate text embeddings $\{w_j\}_{j=1}^K$ for the $K$ training classes, using the prompts $\{\textit{``A \{domain\} of a \{class j\}''}\}_{j=1}^K$. For each image in each dataset, its CLIP image embedding $f_i$ is obtained. The pseudo-label weights $y_i$ are then calculated by:
\begin{equation}
    y_i = \left\{\frac{f_i^{\top}w_j}{||f_i||||w_j||}\right\}_{j=1}^K \text{, and} \quad
    \tilde{y}_i = \arg\max_j \sigma(y_i^j) \text{.}
    \label{eq:pseudo_1n2}
\end{equation}
The $j$-th element of $y_i$ is the cosine similarity score between the image $I_i$ corresponding to the image embedding $f_i$ and the class text embedding $w_j$. The set of vectors $Y^A=\{y_i^A\}_{i=1}^{N} \in \mathbb{R}^{N\times K}$ and $Y^B=\{y_j^B\}_{j=1}^{M} \in \mathbb{R}^{M\times K}$ thus forms the weights of the pseudo-labels,and the pseudo-label class assignment $\tilde{y}_i$ can be obtained by taking the argmax of $\sigma(y_i)$, where $\sigma(\cdot)$ is the Softmax operator.

\subsection{Pseudo-Label Refinement}
To refine the quality of pseudo-labels, we alternate training between updating the parameters of $E_{\theta_q}$ and the pseudo-label weights \cite{Abdelfattah2023}. Every epoch end, we fix our network and update $Y^A$ and $Y^B$ for $n_r$ steps using the Kullback-Leiber (KL) loss function. Specifically, we apply the gradients for every $y_i\in Y$, \ie:

\begin{equation}
    y_i \leftarrow y_i-\psi(y_i) \circ \nabla_{y_i}\operatorname{KL}(\sigma[\hat{x}_i] \: \mid  \: \sigma[y_i] ),
    \label{eq:refine 1}
\end{equation}
\begin{equation}
    \text{where}\quad \hat{x}_i = \left\{\frac{\tilde{x}_i^{\top}w_j}{||\tilde{x}_i||||w_j||}\right\}_{j=1}^K \text{, and}\quad
    \psi(y_i)=\left\{\frac{c_\psi}{\lambda_\psi\sqrt{2\pi}} \exp{\left[-\frac{(y_{ij}-0.5)^2}{2\lambda_{\psi}^2}\right]} \right\}_{j=1}^K \text{.}
    \label{eq:refine 2}
\end{equation}

We drop the domain superscripts in this section for brevity. In this loss function, $\sigma[\hat{x}_i]$ and $\sigma[y_i]$ are the target and predicted distributions respectively. $\hat{x}$ is obtained by a cosine similarity operation with the features $\tilde{x}_i$ in a memory bank encoded by $E_{\theta_k}$ and text embeddings $w_j$ in a similar process used to initialize $y_i$. $\psi(y_i)$ is a normal distribution function of $y_i$ centered at 0.5, used to scale the loss gradients and maximize updating of the least-confident parameters, \ie, those with values around 0.5. $c_\psi$ and $\lambda_{\psi}$ are hyperparameters that correspond to the weight and the standard deviation of $\psi(y_i)$ respectively.

Subsequently, we fix $Y^A$ and $Y^B$ to use as supervision for training $E_{\theta_q}$. For symmetry, we include $\mathcal{L}_{KL}$, which flips the inputs of the KL loss. This alternation of model training and pseudo-label refinement continues until convergence or upon reaching the maximum number of epochs. 

\subsection{Multi-Granular Contrastive Losses}
We propose a multi-granular contrastive learning approach to align features on three levels of granularity using the pseudo-labels as supervision. On an instance-level, we apply the standard contrastive loss from \cite{Chen2020a}: 
\begin{equation}
    \mathcal{L}_{II} = \frac{-1}{|\textbf{I}|}\sum\limits_{i\in\textbf{I}} \log\biggr[\frac{\exp{(x_i^\top x_i'/\tau)}}{\sum_{a\in\mathbf{I}}\exp{(x_i^\top \tilde{x}_a/\tau)}}\biggr],
    \label{eq:II loss}
\end{equation}

\noindent
where $x_i$ and $x'_i$ are features vectors of different augmentations of $I_i$ encoded by $E_{\theta_q}$, while $\tilde{x}_a$ represents feature vector in the memory bank encoded by $E_{\theta_k}$. $\tau$ is an adjustable temperature hyperparameter. We denote $\mathbf{I}$ as indices of the entire training dataset for one domain, and $|\mathbf{I}|$ as its cardinality. 

To ensure that our model learns more semantically meaningful features, we extend contrastive loss to the cluster level. We form positive pairs between instances of the same cluster within a domain, where each cluster contains samples of the same pseudo-labels. We define the inter-cluster loss on domain A as: 
\begin{subequations}
\begin{equation}
    \mathcal{L}_{IC}^A = \frac{-1}{|\textbf{I}^A|}\sum\limits_{i\in\textbf{I}^A} \frac{1}{|P^A(i)|} C^A(x_i^A, \tilde{x}_p^A, \tilde{x}_a^A),
    \label{eq:IC loss}
\end{equation}
\vspace{-2mm}
\text{where}
\vspace{-2mm}
\begin{equation}
C^A = \sum\limits_{p\in P^A(i)} \log\biggr[\frac{\exp{(x_i^{A,\top} \tilde{x}_p^A/\tau)}}{\sum_{a\in\mathbf{I}^A}\exp{(x_i^{A,\top} \tilde{x}_a^A/\tau)}}\biggr].
    \label{eq:IC loss2}
\end{equation}
\end{subequations}
To help bridge the domain gap, we further extend the loss function to the domain level. We form positive pairs between instances of the same cluster across domains. We define the inter-domain loss on domain A as:
\begin{subequations}
\begin{equation}
    \mathcal{L}_{ID}^A = \frac{-1}{|\textbf{I}^A|}\sum\limits_{i\in\textbf{I}^A} \frac{1}{|P^B(i)|} D^A(x_i^A, \tilde{x}_p^B, \tilde{x}_a^B)
    \label{eq:ID loss}
\end{equation}
\vspace{-2mm}
\text{where}
\vspace{-2mm}
\begin{equation}
    D^A = \sum\limits_{p\in P^B(i)} \log\biggr[\frac{\exp{(x_i^{A,\top} \tilde{x}^B_p/\tau)}}{\sum_{a\in\mathbf{I}^B}\exp{(x_i^{A,\top} \tilde{x}^B_a/\tau)}}\biggr] \text{.}
    \label{eq:ID loss2}
\end{equation}
\end{subequations}

\noindent
Here, $P(i)$ represents the indices for the set of samples belonging to the same assigned pseudo-class as $I_i$, and $|P(i)|$ as its cardinality. Specifically, for an image $I_{i}^A$ from domain A, $P^A(i) = \{p\in \mathbf{I}^A:\tilde{y}_p^A=\tilde{y}_i^A\}$ and $P^B(i) = \{p\in \mathbf{I}^B:\tilde{y}_p^B=\tilde{y}_i^A\}$. $\mathcal{L}_{IC}^B$ and $\mathcal{L}_{ID}^B$ can be calculated in the same way.

For every loss objective, we average the loss between domains, \ie $\mathcal{L} = 0.5(\mathcal{L}^A + \mathcal{L}^B)$, thus forming our full training objective: 
\begin{equation}
\mathcal{L}_{total} = w_{KL}\mathcal{L}_{KL} + w_{II}\mathcal{L}_{II} + w_{IC}\mathcal{L}_{IC} + w_{ID}\mathcal{L}_{ID},    
\end{equation}
where $w_{KL}$, $w_{II}$, $w_{IC}$, and $w_{ID}$ are hyperparameters to balance the loss terms.

\subsection{Cross-Domain Mapping Function} \label{mappingfunction}
While paired cross-domain images are not available, it is trivial to generate paired text embeddings using prompt templates. We make the assumption that embeddings of domain-specific prompt templates are a good approximation for their image counterparts, and therefore a mapping function obtained from the text embeddings would work on the image embeddings:
\begin{equation}
    \Omega w^A = w^B \implies \Omega x^A = x^B,
    \label{eq:omega1}
\end{equation}
where $\Omega$ is a transformation matrix that maps features from one domain to another. Paired text embeddings $w^A$ and $w^B$ can be obtained from encoding the text template \textit{``A \{domain\} of \{object\}''} using text encoder $E_{t}$. For example, \textit{``An infograph of an airplane''} is paired with \textit{``A photo of an airplane''}. Using internet-sourced object names, we create 3,000 text embeddings pairs for each domain pair. If $\Omega$ is assumed to be orthogonal, then the optimal solution $\text{argmin}_{\Omega} \| \Omega w^A -w^B\|^2$ can be solved in closed-form: 

\begin{equation}
    \Omega=UV^{\top} \text{,}
    \label{eq:omega2}
\end{equation}
where $U\Sigma V^{\top}$ is the decomposition of $w^Bw^{A\top}$ \cite{Schoenemann1966}. 
$\Omega$ can be directly inserted into our model even before training to improve performance.
We found empirically that allowing the parameters of $\Omega$ update along with the rest of the model during training yields better results. It should be noted that the cross-domain mapping function improves cross-domain alignment without being adversely affected by noisy pseudo labels since it is learned fully from text embeddings.

\subsection{Prompt Learning}
To handle unseen categories and prevent \textit{catastrophic forgetting}, we follow \cite{Sain2023} to freeze all layers of our CLIP-pretrained domain alignment encoder except for the Layer Norm layers. 
A set of learnable prompts $\text{v}\in\mathbb{R}^{n_p\times d_p}$ is then concatenated with image features from both domains for knowledge distillation and preserving the generalization ability of CLIP.
$n_p$ and $d_p$ are hyperparameters that correspond to the number and dimensionality of prompts respectively.

The remaining trainable parameters of $E_{\theta_q}$ and $\text{v}$ are updated using the loss functions described in the above sections. Limiting the training to a smaller subset of features has the added benefit of expediting convergence, leading to reduced overall training time. 

In summary, we integrate CLIP extensively into our framework, encompassing pseudo-label initialization, multi-granular contrastive losses, the approximation of a no-train domain-aligning mapping function, and prompt learning. Detailed descriptions of our hyperparameter tuning strategy can be found in the supplementary material.

\section{Experiments}
\label{sec:experiments}

\subsection{Dataset}
We evaluate on the established ZS-SBIR datasets \textbf{Sketchy Extended \cite{liu2017}}, \textbf{TUBerlin Extended \cite{zhang2016sketchnet}} and \textbf{QuickDraw Extended \cite{Dey2019}}, and follow \cite{Yelamarthi2018} by reporting Precision scores for Sketch to Image domain. Additionally, we propose \textbf{DomainNet Zero-Shot}, a subset of DomainNet \cite{Peng2019} which spans across 6 domains (Clipart (C), Infograph (I), Sketch (S), Quickdraw (Q), Painting (P) and Real (R)) and shares the same 345 categories as the original Quickdraw dataset \cite{Ha2017}. Thus, we follow \cite{Dey2019} and split DomainNet into the same train/test split, resulting in over 200,000 samples across 6 domains. Following \cite{Hu2022}, we report results on 6 domain pairings: C-S, I-R, I-S, P-C, P-Q and Q-R, ensuring that every domain is experimented on twice, and retrieval is evaluated bidirectionally (\ie C$\rightarrow$S and S$\rightarrow$C). We plot the distribution of images across classes and domains for visualization in the supplementary material. Finally, we randomly sample approximately 12 samples per training category for every dataset, which we set aside as a validation dataset.

\subsection{Implementation Details}
Our proposed approach is implemented in PyTorch on a 24GB NVIDIA RTX 4090 GPU. For both initializing the pseudo-labels and warm-starting our backbone encoder, we use the ViT encoder of CLIP with ViT-B/32 weights. Only the prompts and the weights of Layer Normalization layers in $E_{\theta_q}$ are set to update in training. These parameters are trained with a SGD optimizer with learning rate set to $0.0002$, applied with a cosine learning rate schedule that gradually reduces it to 0, and with early stopping employed for regularization. Following \cite{Sain2023}, we inject our prompts $\text{v}\in\mathbb{R}^{10\times768}$ into the first layer of the transformer, and use the same encoder and prompts for both domain inputs. Resizing all images to $224 \times 224$ pixels, we train our model with a batch size of 64 for 25 epochs, and it maps the input image into a feature space $x\in\mathbb{R}^{64}$. With the reduced learnable parameters, we observed quick training convergence: 9.5 GB RAM and 21 minutes across all DomainNet Zero-Shot pairings, and 2.1 GB RAM and 25 seconds for inference.

\newcommand\y{1.2cm}

\begin{table}[!t]
    \centering
    \begin{minipage}[t]{0.485\textwidth}\vspace{0pt}
        \centering
        \resizebox{\textwidth}{!}{
            \begin{tabular}{p{\y}rccc}
                \toprule
                \multirow{2}[3]{\y}{Setting} & \multirow{2}[3]{*}{Method} & Sketchy & TUBerlin & Quickdraw \\ 
                \cmidrule(l){3-5}            &                            & P@200   & P@100    & P@200 \\
                \midrule \addlinespace
                \multirow{7}[1]{\y}
                    {US ZS} & CDS \cite{Kim2021}               & 24.99 & 18.58 & 6.49  \\
                            & PCS \cite{Yue2021}               & 13.29 & 13.89 & 3.65  \\
                            & CoDA \cite{Wang2023}             & 17.02 & 15.85 & 5.15  \\
                            & CLIP-UCDIR \cite{Hu2022}         & 40.63 & 38.75 & 8.90 \\
                            & Vanilla CLIP \cite{Radford2021a} & 35.01 & 45.40 & 8.42 \\
                            & AMA \cite{yin2024asymmetric}     & 58.50 & 59.20 & 19.20 \\
                \midrule \addlinespace
                \multirow{2}{\y}
                    {WS ZS} & CLAIR-A (Ours-A)                 & 64.64 & 71.17 & 21.49 \\
                            & CLAIR-B (Ours-B)                 & \textbf{65.74} & \textbf{72.11} & \textbf{23.07} \\
            \end{tabular}     
        }
    \end{minipage}
    \hfill
    \begin{minipage}[t]{0.485\textwidth}\vspace{0pt}
        \centering
        \resizebox{\textwidth}{!}{
            \begin{tabular}{p{\y}rccc}
                \toprule
                \multirow{2}[3]{\y}{Setting} & \multirow{2}[3]{*}{Method} & Sketchy & TUBerlin & Quickdraw \\ 
                \cmidrule(l){3-5}            &                            & P@200   & P@100    & P@200 \\
                \midrule \addlinespace
                \multirow{8}[1]{\y}
                    {FS ZS} & TCN \cite{wang2021transferable}   & 60.8 & 61.6 & 29.8 \\
                            & TVT \cite{Tian2022a}              & 61.8 & 66.2 & 29.3 \\
                            & PSKD [ViT] \cite{wang2022prototype} & 64.5 & 66.2 & 29.8 \\
                            & Sketch3T \cite{Sain2022}          & 64.8 & 67.1 & - \\
                            & ZSE-SBIR \cite{Lin2023}           & 62.40 & 65.70 & 21.60 \\
                            & SketchLVM \cite{Sain2023}         & 72.50 & 73.20 & 38.80 \\
                            & Dr.CLIP \cite{li2024dr}           & 70.60 & 76.30 & 31.20 \\
                            & DCDL \cite{li2025dcdl}            & 76.90 & 74.10 & 29.60 \\
            \end{tabular}
        }
    \end{minipage}
    \vspace{2mm} 
    \caption{Experimental results on the Zero-Shot test splits of Sketchy Extended, TUBerlin Extended and QuickDraw Extended datasets. \textbf{Bold} represent best results. \textbf{Unsupervised Zero-Shot (US ZS)}: The model is trained with unlabeled images. \textbf{Weakly Supervised Zero-Shot (WS ZS)}: The model is trained with noisy pseudo labels generated from CLIP. \textbf{Fully Supervised Zero-Shot (WS ZS)}: The model is trained with image labels as supervision. CLAIR-A uses the same hyperparameters across all experiments while CLAIR-B follows AMA to tune hyperparameters on Sketchy. More details on our baseline methods can be found in the supplementary section.}
\label{tab:Performance comparison others v2}
\end{table}

\subsection{Evaluation Metrics}
As our objective involves category-level image retrieval, we adopt the evaluation metrics from \cite{Kim2021, Hu2022} and calculate precision at the top $k$ retrievals (P@$k$) for all samples across all classes. We report P@100 for TUBerlin Extended and P@200 for Sketchy Extended and Quickdraw Extended datasets, aligning with ZS-SBIR literature. For DomainNet Zero-Shot, we report P@50 scores bidirectionally for each domain pair. Specifically, for two domains A and B, scores are reported with A as the query domain and B as the target domain and vice versa. In a similar vein with recent UCDIR works, we provide additional P@100 and P@200 scores, as well as Precision-Recall curves for all DomainNet Zero-Shot experiments in the supplementary material.
\vspace{5mm}
\subsection{Results}

\begin{table*}[!t]
\centering
\resizebox{0.95\textwidth}{!}{

\begin{tabular}{rccccccccccccc}
\toprule

\multicolumn{14}{c}{DomainNet Zero-Shot} \\
\cmidrule(l){2-14}

Method & C$\rightarrow$S & S$\rightarrow$C & I$\rightarrow$R & R$\rightarrow$I & I$\rightarrow$S & S$\rightarrow$I & P$\rightarrow$C & C$\rightarrow$P & P$\rightarrow$Q & Q$\rightarrow$P & Q$\rightarrow$R & R$\rightarrow$Q & Avg. \\

\midrule \addlinespace
CDS \cite{Kim2021}               & 36.64 & 39.07 & 38.26 & 24.61 & 24.51 & 18.27 & 37.87 & 39.68 & 8.27 & 11.28 & 12.60 & 8.14 & 24.93    \\
PCS \cite{Yue2021}               & 34.37 & 30.71 & 17.85 & 34.37 & 17.99 & 17.59 & 35.72 & 26.35 & 8.38 & 6.55  & 13.82 & 11.95 & 21.30  \\
CoDA \cite{Wang2023}             & 30.65 & 32.85 & 35.31 & 18.22 & 19.67 & 15.06 & 29.37 & 33.30 & 5.99 & 7.22  & 7.99  & 6.49 & 20.18  \\
CLIP-UCDIR \cite{Hu2022}         & 42.43 & 41.06 & 21.02 & 24.73 & 28.35 & 35.02 & 43.25 & 42.32 & 9.31 & 10.27 & 10.68  & 9.78 & 26.52  \\
Vanilla CLIP \cite{Radford2021a} & 67.07 & 63.57 & 56.62 & 41.24 & 49.13 & 28.50 & 61.00 & 51.44 & 34.45 & 9.50  & 8.41  & 38.39 & 42.44 \\
CLAIR-A (Ours)             & \textbf{81.61} & \textbf{76.24} & \textbf{56.79} & \textbf{68.92} & \textbf{51.53} & \textbf{60.34} & \textbf{74.73} & \textbf{81.19} & \textbf{44.93} & \textbf{26.24} & \textbf{26.32} & \textbf{51.14} & \textbf{58.33} \\

\bottomrule
\end{tabular}

}
\vspace{1.5mm}
\caption{
Precision at top 50 (\%) on the Zero-shot test splits of all domain pairings in the DomainNet Zero-Shot dataset, compared against Unsupervised baselines. Domains: C - Clipart, S - Sketch, I - Infograph, P - Painting, Q - Quickdraw, R - Real. 
}
\vspace{-2mm}
\label{tab:Performance comparison domainnet condensed}
\end{table*}

\Cref{tab:Performance comparison others v2} shows that our proposed CLAIR outperforms AMA on all comparable datasets, thus validating our methodology. Additionally, \cref{tab:Performance comparison domainnet condensed} demonstrates the generalizability of our CLAIR, surpassing all Unsupervised baselines on every domain pair from the DomainNet Zero-Shot dataset. Notably, both our CLAIR and CLIP-UCDIR initiate training with identical weights inherited from Vanilla CLIP. The performance of CLIP-UCDIR declines after fine-tuning, reminiscent of catastrophic forgetting. In comparison, our model improves on Vanilla CLIP by 15.89\% on average in terms of P@50 on all datasets. Some qualitative results (including failure cases) are illustrated in the supplementary material. We observe that most of the top images, even the false positives, retrieved by our method resemble the query image in likeliness. In fact, several false positives are due to images containing multiple objects (\eg a monkey on a tree) but are only assigned one class, illustrating the limitation of the single-label nature of the dataset. 

An interesting observation arises when considering the performance of CLIP in scenarios where Quickdraw serves as the query domain. Consisting of a large amount of rudimentary drawn doodles, the high level of abstraction presents a significant domain gap. Consequently, CLIP faces challenges in such contexts. In comparison, our CLAIR not only mitigates these challenges but also demonstrates substantial improvement over CLIP in such domain pairings by an average P@50 score of 14.65\%, 16.7\% and 17.9\% on the QuickDraw extended dataset, and the Q $\rightarrow$ P and Q $\rightarrow$ R pairings of the DomainNet Zero-Shot dataset, respectively. This performance differential signifies the robustness of our training approach in addressing domain-specific intricacies and optimizing retrieval accuracy.  

\subsection{Ablation Study}

\begin{table}[!t]
    \centering
    \begin{minipage}[t]{0.445\textwidth}\vspace{0pt}
        \centering
        \resizebox{\textwidth}{!}{
            \begin{tabular}{rcccc}
                \toprule
                Method & C $\rightarrow$ S & S $\rightarrow$ C & I $\rightarrow$ R & R $\rightarrow$ I \\
                \cmidrule(l){2-5}
                w/o mapping & 67.07 & 63.57 & 56.61 & 41.24 \\
                w/ mapping & \textbf{70.36} & \textbf{69.46} & \textbf{59.48} & \textbf{55.81} \\
                \midrule
                Method & I $\rightarrow$ S & S $\rightarrow$ I & P $\rightarrow$ C & C $\rightarrow$ P \\
                \cmidrule(l){2-5}
                w/o mapping & 49.13 & 28.50 & 61.00 & 51.44 \\
                w/ mapping & \textbf{52.67} & \textbf{42.61} & \textbf{66.26} & \textbf{69.73} \\                
                \midrule
                Method & P $\rightarrow$ Q & Q $\rightarrow$ P & Q $\rightarrow$ R & R $\rightarrow$ Q \\         
                \cmidrule(l){2-5}
                w/o mapping & 34.45 & 9.50 & 8.42 & 38.39 \\
                w/ mapping & \textbf{36.71} & \textbf{12.87} & \textbf{10.78} & \textbf{40.48} \\
                \midrule
                \multirow{1.5}[1]{*}{Average P@50} & \multicolumn{2}{c}{Without mapping} & \multicolumn{2}{c}{With mapping} \\
                \cmidrule(l){2-5} & \multicolumn{2}{c}{42.44} & \multicolumn{2}{c}{\textbf{48.93}} \\
                \bottomrule
            \end{tabular}
        }
    \end{minipage}%
    \hfill%
    \begin{minipage}[t]{0.525\textwidth}\vspace{0pt}
        \centering
        \resizebox{\textwidth}{!}{
            \newcommand\x{3.25cm}
            \begin{tabular}{p{\x}lc}
                \toprule
                Ablation & Experiment & Avg. P@50 \\
                \midrule
                & Full model & \textbf{57.52} \\
                \midrule
                \multirow{8}[3]{\x}{(i) Leave one out}
                & w/o prompt & 53.40 \\
                & w/o mapping & 54.68 \\
                & w/o refinement & 51.62 \\
                & w/o freeze & 50.64 \\
                & w/o $\mathcal{L}_{KL}$ & 56.51 \\
                & w/o $\mathcal{L}_{II}$ & 55.44 \\
                & w/o $\mathcal{L}_{IC}$ & 54.50 \\
                & w/o $\mathcal{L}_{ID}$ & 50.11 \\
                \midrule
                \multirow{2}[1]{\x}{(ii) Expanded class list}
                & Full expanded list & 55.24 \\
                & Exclude training class names & 54.56 \\
                \midrule
                \multirow{1}[1]{\x}{(iii) In the wild dataset}
                & Mismatched categories & 57.09 \\
                \bottomrule
            \end{tabular}
        }
    \end{minipage}
    \vspace{3mm}
    \caption{
        [Left] Effect of cross-domain mapping function on image retrieval using Vanilla CLIP. [Right] The combined table of three ablation studies: (i) significance of each component, (ii) effect of using training class names to initialize pseudo-labels, (iii) impact of having mismatched classes between domains. P@50 scores are averaged across all domain pairings from the DomainNet Zero-Shot dataset.
    }
    \label{tab:Combined ablation}
\end{table}

\paragraph{Influence of the Cross-Domain Mapping Function.} \Cref{tab:Combined ablation} shows the impact of the cross-domain mapping function $\Omega$. The inclusion of $\Omega$ improves performance of Vanilla CLIP on all DomainNet Zero-Shot pairs by an average of $6.49\%$.
This affirms that a mapping function derived solely from domain-specific text prompts without exposure to any training images enhances retrieval scores within their respective image domains, and aligns with our objective of initializing the model with a closed-form solution to mitigate the domain gap prior to training.

\vspace{-3.5mm}\paragraph{Performance contribution of components.} \Cref{tab:Combined ablation} accesses the impact of each module on the performance of our model by systematically excluding one component at a time. The results validate our design choices in the model architecture, where every component contributed to its performance. Moreover, it also shows the removal of a single component does not make training fail catastrophically. Of all the components, the most significant ones are the inter-domain contrastive loss and the freezing of CLIP weights, elevating P@50 accuracy from 50.11\% and 50.64\% to 57.52\%, respectively.

\vspace{-3.5mm}\paragraph{Impact of using training class names.} CLAIR makes use of the training category names to initialize pseudo-labels. To assess the potential bias introduced by using the category names, we propose an alternate source and generate the pseudo labels using the 3000 object names used to solve the cross-domain mapping function. To ensure a zero-shot setting, the test category names are excluded from the expanded list. Additionally, we further exclude all training classes from this expanded list to evaluate the impact of its inclusion during training. \Cref{tab:Combined ablation} shows that having the names of training classes is helpful but not essential. Expanding the pseudo-label list of classes and excluding the training labels specifically only leads to a performance degradation of 2.28\% and 2.96\% respectively. Notably, these scores still outperform all other unsupervised baselines.

\vspace{-3.5mm}\paragraph{Assessment on mismatched categories.} The datasets used in our evaluation all share the same classes across domains. This assumption may not hold for more realistic, or `in-the-wild' datasets, especially for unsupervised tasks. To assess this impact on our proposed method, we further subset the train split of the DomainNet Zero-Shot dataset into two groups, with 60 classes in each group. Each domain consists of 20 unique classes not found in the other domain, with the remaining 40 classes shared. Naturally, these class distributions are not known by our model during training. We observe from \cref{tab:Combined ablation} that performance degradation is almost negligible (0.43\%), which means that our model can perform well even if the categories in the query and target domains are mismatched.

\section{Conclusion}
\label{sec:conclusion}
In this paper, we propose the problem setting of Weakly Supervised Zero-Shot Cross-Domain Image Retrieval (WSZS-CDIR) to alleviate the burden of human annotation and enable retrieval on any novel categories, and introduce CLAIR to solve this task. Our CLAIR first uses CLIP to generate noisy pseudo-labels which we then refine iteratively alongside a domain-aligned feature encoder throughout the training process. To enhance cross-domain image alignment, we propose a novel contrastive learning strategy that strategically selects positive and negative pairs at various levels of granularity, ranging from inter-instance samples to inter-domain clusters. We also design an effective cross-domain mapping function that is unaffected by noisy pseudo-labels to further improve the alignment of features across domains. Experimental results on the Sketchy Extended, TUBerlin Extended, QuickDraw Extended and DomainNet Zero-Shot datasets demonstrate the superiority of our CLAIR over baseline approaches for every domain pairing.

\bibliography{egbib}

\newpage
\appendix
\section{DomainNet Zero-Shot Dataset}
\begin{figure*}[h]
    \centering
    \includegraphics[width=\textwidth]{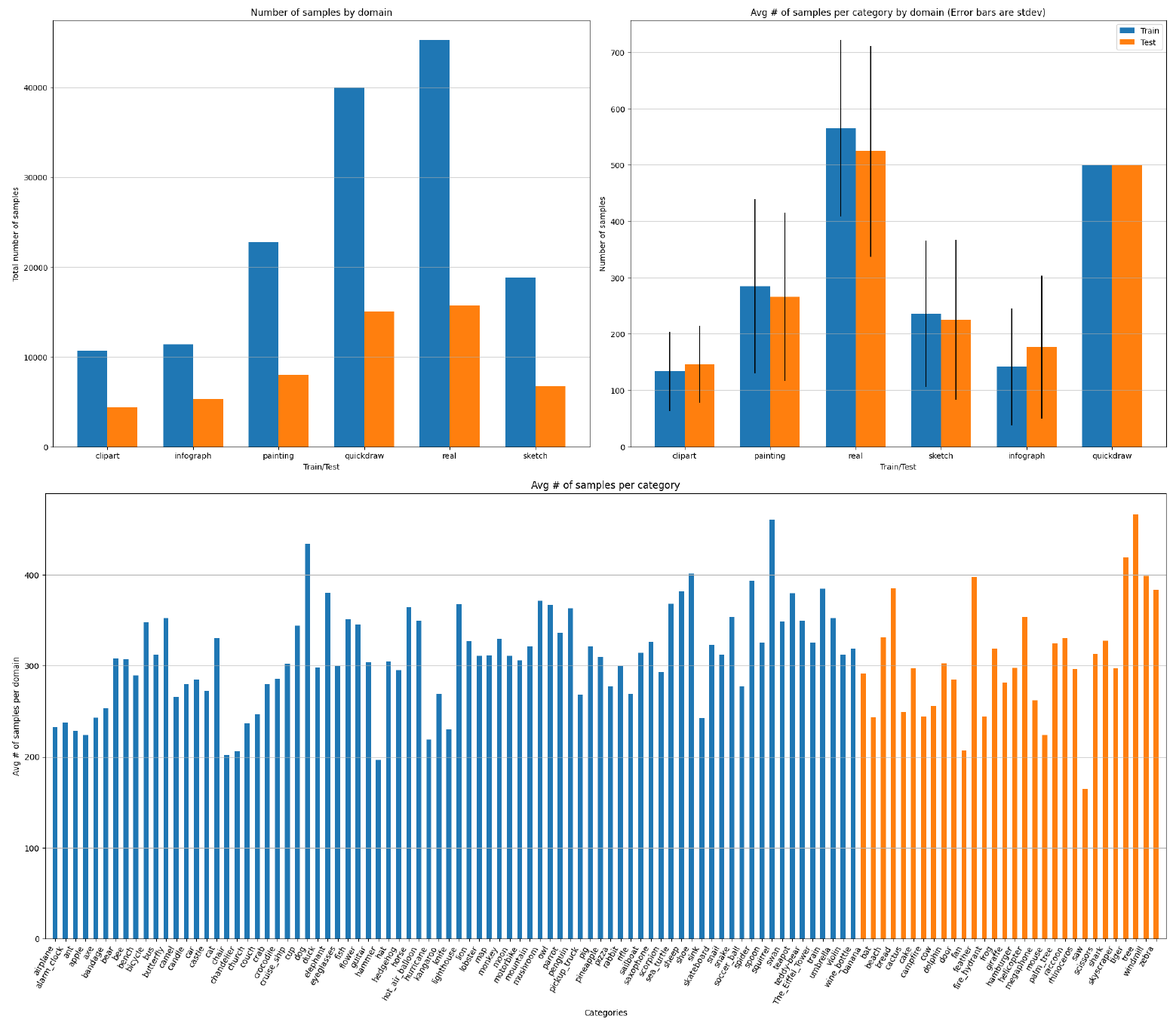}
     \caption{DomainNet Zero-Shot statistics. [Top-left] Total number of samples by domain, split into train and test sets. [Top-right] Average number of samples per label by domain, split into train and test sets. Vertical error bars correspond to standard deviation of samples of all labels. [Bottom] Average number of samples per domain by label, colorized into train and test sets.}
  \label{fig:dataset stats}
\end{figure*}

To evaluate our proposed approach on across various domain pairings, we establish the DomainNet Zero-Shot dataset by re-splitting the original DomainNet dataset \cite{Peng2019} such that the test set categories do not overlap with the train set. Our finalized dataset consists of 148,998 training samples and 55,169 testing samples, split into 80 and 30 classes, respectively. \cref{fig:dataset stats} provides a detailed breakdown of the dataset.

\section{Model and Hyperparameter selection}
The model parameters are trained with a SGD optimizer with learning rate set to $0.0002$, applied with a cosine learning rate schedule that gradually reduces it to 0, and with early stopping employed for regularization. Following \cite{Sain2023}, we inject our prompts $\text{v}\in\mathbb{R}^{10\times768}$ into the first layer of the transformer, and use the same encoder and prompts for both domain inputs. Resizing all images to $224 \times 224$ pixels, we train our model with a batch size of 64 for 25 epochs, and it maps the input image into a feature space $x\in\mathbb{R}^{64}$. 

To adhere to the zero-shot setting, we perform hyperparameter and model selection using the validation dataset. This set was curated by randomly sampling approximately 12 images per category per domain, which was withheld from training. Validation set metrics were used as conditions for early stopping, and subsequently for best performing model selection for each experiment. Similarly, hyperparameters were optimized using a grid search based on the P@50 scores from the validation set. CLAIR-A was tuned on the Clipart-Sketch pair from the DomainNet Zero-Shot dataset and evaluated on all Zero-Shot test sets. Following AMA, CLAIR-B was tuned on the Sketchy dataset and tested on the Zero-Shot test sets of Sketchy, TUBerlin and Quickdraw. The specific parameter values for $m$, $w_{KL}$, $w_{II}$, $w_{IC}$, $w_{ID}$, $c_\psi$ and $\lambda_{\psi}$ in CLAIR-A, CLAIR-B, and the ablation study are ($0.9$, $0.5$, $0.5$, $1.0$, $1E4$, $20$), ($0.9$, $1E2$, $1.0$, $0.5$, $0.5$, $1E4$, $20$) and ($0.9$, $1.0$, $1.0$, $1.0$, $1.0$, $1E4$, $20$) respectively.

\section{Baselines}
We propose the following Unsupervised baselines and re-train on our datasets to provide comparison to our approach. \textbf{CDS} \cite{Kim2021} is an Unsupervised Cross-Domain Pre-Training algorithm. It uses a combination of instance discrimination and cross-domain matching to learn a common embedding space. \textbf{PCS} \cite{Yue2021} is also an Unsupervised learning approach for Domain Adaptation. It learns a common embedding space by a combination of in-domain and cross-domain prototypical contrastive self-supervision. \textbf{CoDA} \cite{Wang2023} adopts self-matching supervision and cross-domain classifier alignment learning mechanisms to perform UCDIR. We acknowledge the extensive data used to pre-train CLIP may provide a potential advantage when comparing to non-CLIP methods. To mitigate this issue, we replace the image encoder in \cite{Hu2022}, which was specifically designed for UCDIR, with the identical pre-trained CLIP image encoder employed in our experiments. This modified model, termed \textbf{CLIP-UCDIR}, is then retrained under the same conditions as outlined in \cite{Hu2022}. We also compare directly with the visual embedding space of \textbf{Vanilla CLIP} \cite{Radford2021a} without modifications or additional fine-tuning. The latter two baselines offer a performance comparison with our solution under scenarios where it undergoes training without accounting for zero-shot supervision, or no training at all.

\textbf{AMA} \cite{yin2024asymmetric} is designed for UZS-SBIR, using an asymmetric learning strategy that incorporates self-distillation and guided clustering, and bears the most similarity to our work. However, no code has been released at this time. Hence, we only provide comparison on overlapping sketch-based datasets. We further extend this comparison with recent ZS-SBIR approaches, which are trained under the supervised setting, \ie images are fully labelled during training. Notably, \textbf{SketchLVM} \cite{Sain2023}, \textbf{Dr.CLIP} \cite{li2024dr} and \textbf{DCDL} \cite{li2025dcdl} are all CLIP-based methods. Despite this, our method produces results that are within range of the state-of-the-art baselines. 

\section{Qualitative Results on DomainNet Zero-Shot}
\begin{figure*}[h]
    \centering
    \includegraphics[width=\textwidth]{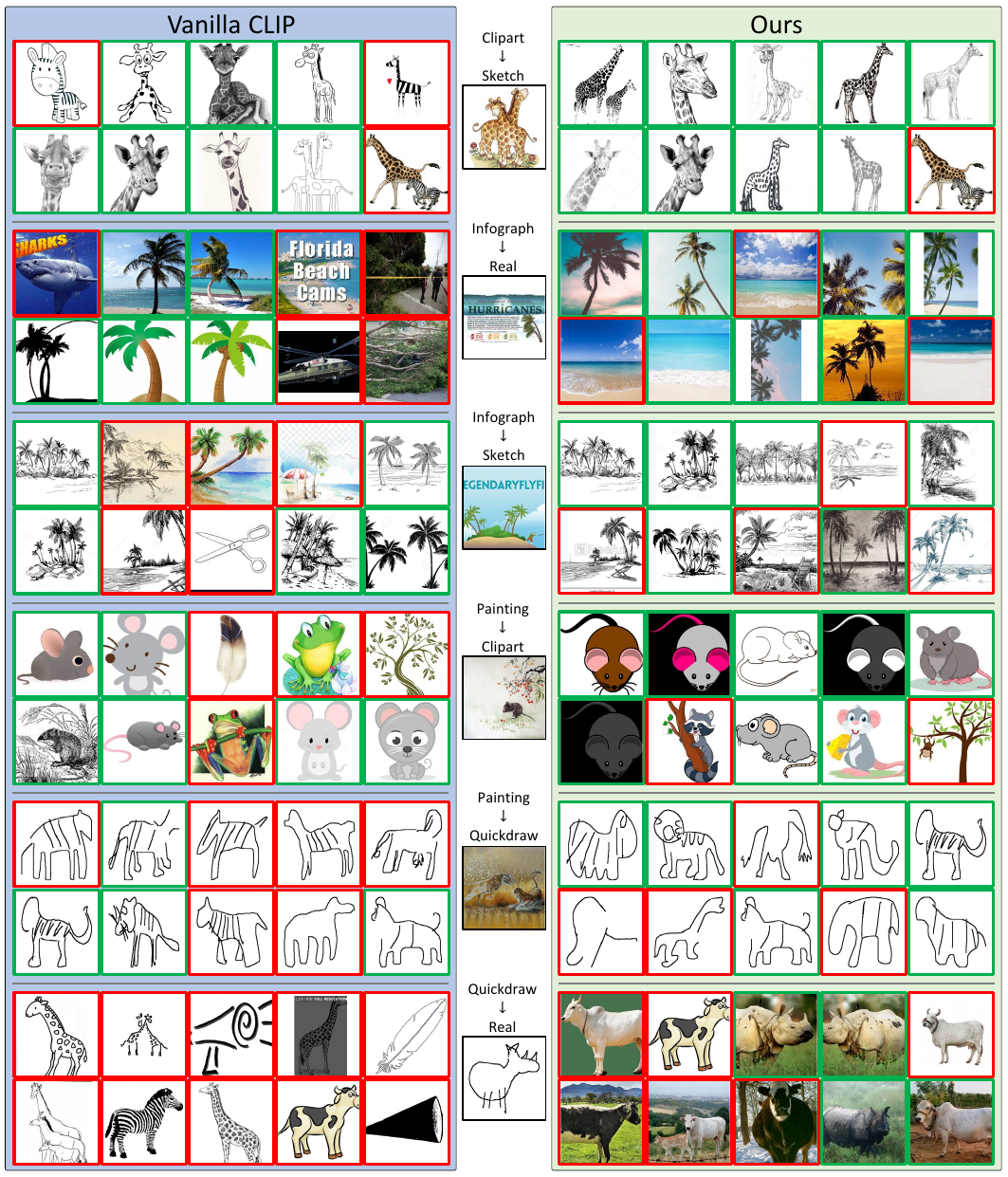}
     \caption{A qualitative assessment showcasing the top 10 retrieval outcomes for queries from six domain pairs. The images were randomly selected from the test set, and correct and incorrect retrievals are marked with green and red outlines, respectively. The assigned categories for the query images, from top to bottom, are Giraffe, Palm Tree, Palm Tree, Mouse, Tiger, and Rhinoceros. }
  \label{fig:qualitative}
\end{figure*}

We showcase retrieval examples of our model against CLIP in \cref{fig:qualitative}, highlighting the evident improvement our model achieves in image retrievals over CLIP within the test set. Observing the images rationalizes some instances of incorrect retrievals by both models. For instance, an image annotated as the ``zebra'' class was correctly retrieved by both models for a query image with the ``giraffe'' label, as it contains both animals. In another instance, the query image of class ``mouse'' also contains a tree, resulting in some images corresponding to ``tree'' labels being retrieved.

\section{Full Quantitative Results}
\begin{figure*}[h]
    \centering
    \includegraphics[width=\textwidth]{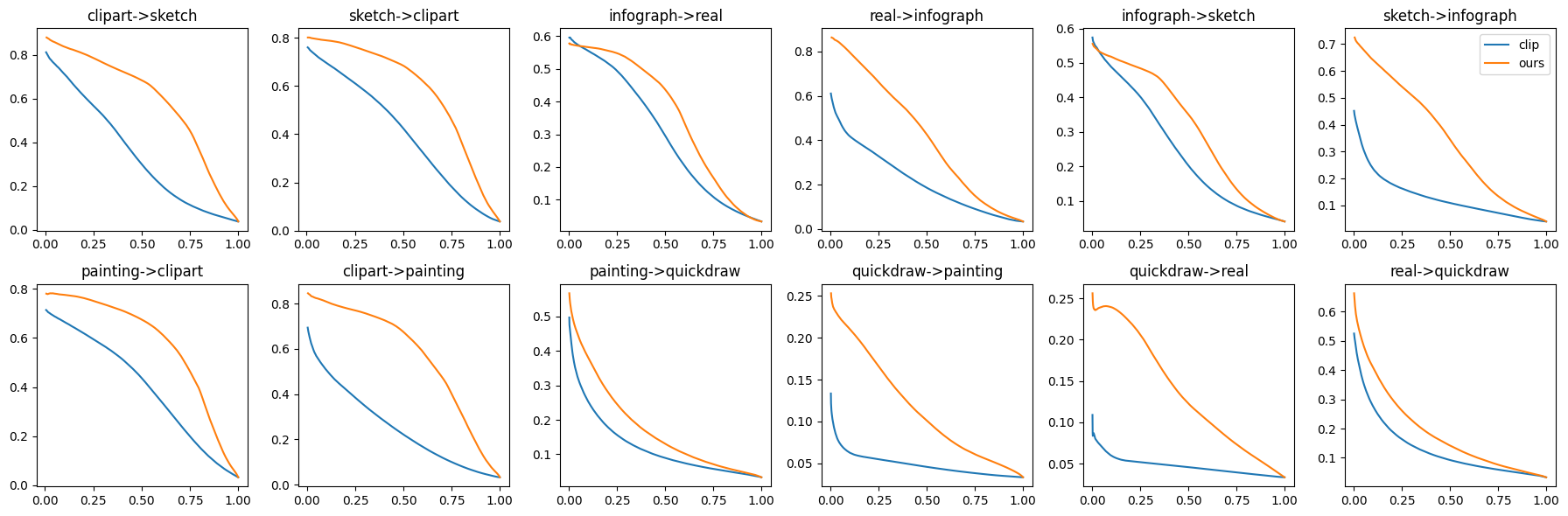}
    \vspace{-2mm}
     \caption{Precision Recall curves of every domain pair from the DomainNet Zero-Shot dataset.}
  \label{fig:prcurve}
\end{figure*}

In line with the evaluation methodology of UCDIR tasks, we present the P@50, P@100 and P@200 scores from all four datasets in \cref{tab:Performance comparison domainnet full} and \cref{tab:Performance comparison others v1}. To provide a better understanding of our performance, we also plot the Precision-Recall (PR) curve for every domain pairing from the DomainNet Zero-Shot dataset in \cref{fig:prcurve}. We can observe that our proposed approach outperforms Vanilla CLIP at all thresholds.

\vspace{5mm}

\begin{table*}[h]
\centering
\resizebox{1.0\textwidth}{!}{

\begin{tabular}{rccccccccc}
\toprule

\multirow{2}[3]{*}{Method} & \multicolumn{3}{c}{Sketchy Extended} & \multicolumn{3}{c}{TUBerlin Extended} & \multicolumn{3}{c}{Quickdraw Extended} \\ 
\cmidrule(l){2-10}         & P@50 & P@100 & P@200 & P@50 & P@100 & P@200 & P@50 & P@100 & P@200 \\

\midrule \addlinespace
CDS \cite{Kim2021}               & 30.76 & 28.29 & 24.99 & 22.00 & 18.58 & 18.58 & 6.61 & 6.49 & 6.49  \\
PCS \cite{Yue2021}               & 15.84 & 14.72 & 13.29 & 15.26 & 13.89 & 12.77 & 3.67 & 3.62 & 3.65  \\
CoDA \cite{Wang2023}             & 21.21 & 19.06 & 17.02 & 18.07 & 15.85 & 15.85 & 5.61 & 5.37 & 5.15  \\
CLIP-UCDIR \cite{Hu2022}         & 48.33 & 45.01 & 40.63 & 42.23 & 38.75 & 35.28 & 9.85 & 9.39 & 8.90 \\
Vanilla CLIP \cite{Radford2021a} & 39.17 & 37.79 & 35.01 & 44.91 & 45.40 & 44.72 & 8.80 & 8.53 & 8.42 \\
CLAIR-A                          & 66.46 & 65.97 & 64.64 & 72.58 & 71.17 & 69.02 & 19.48 & 22.26 & 21.49 \\
CLAIR-B                          & \textbf{69.23} & \textbf{68.15} & \textbf{65.74} & \textbf{73.90} & \textbf{72.11} & \textbf{69.36} & \textbf{22.37} & \textbf{24.74} & \textbf{23.07} \\

\bottomrule
\end{tabular}
 }
\vspace{2mm}
\caption{
Zero-shot Unsupervised Cross-domain Retrieval Accuracy (\%) on test splits of Sketchy Extended, TUBerlin Extended and QuickDraw Extended datasets. Bolded numbers represent best results.
}
\label{tab:Performance comparison others v1}
\end{table*}

\begin{table*}[h]
\centering
\resizebox{\textwidth}{!}{

\begin{tabular}{rccccccccc}
\toprule



 & \multicolumn{9}{c}{DomainNet Zero-Shot} \\
\cmidrule(l){2-10}

\multirow{2}[3]{*}{Method} & \multicolumn{3}{c}{Clipart $\rightarrow$ Sketch} & \multicolumn{3}{c}{Sketch $\rightarrow$ Clipart} & \multicolumn{3}{c}{Infograph $\rightarrow$ Real} \\ 
\cmidrule(l){2-10}         & P@50 & P@100 & P@200 & P@50 & P@100 & P@200 & P@50 & P@100 & P@200 \\

\midrule \addlinespace
CDS \cite{Kim2021}               & 36.64 & 33.12 & 29.57 & 39.07 & 36.19 & 32.89 & 38.26 & 32.02 & 26.46 \\
PCS \cite{Yue2021}               & 34.37 & 31.66 & 28.77 & 30.71 & 27.71 & 25.27 & 17.85 & 17.02 & 16.05 \\
CoDA \cite{Wang2023}             & 30.65 & 27.69 & 24.99 & 32.85 & 29.97 & 26.88 & 35.31 & 29.07 & 24.08 \\
CLIP-UCDIR \cite{Hu2022}         & 42.43 & 39.31 & 36.31 & 41.06 & 36.84 & 33.33 & 21.02 & 19.85 & 18.41 \\
Vanilla CLIP \cite{Radford2021a} & 67.07 & 60.32 & 51.55 & 63.57 & 57.72 & 50.16 & 56.62 & 54.54 & 50.66 \\
CLAIR-A                          & \textbf{81.61} & \textbf{78.75} & \textbf{74.98} & \textbf{76.24} & \textbf{73.57} & \textbf{67.92} &  \textbf{56.79} & \textbf{56.31} & \textbf{55.07}  \\
\addlinespace \midrule

\multirow{2}[3]{*}{Method} & \multicolumn{3}{c}{Real $\rightarrow$ Infograph} & \multicolumn{3}{c}{Infograph $\rightarrow$ Sketch} & \multicolumn{3}{c}{Sketch $\rightarrow$ Infograph} \\
\cmidrule(l){2-10}         & P@50 & P@100 & P@200 & P@50 & P@100 & P@200 & P@50 & P@100 & P@200 \\

\midrule \addlinespace
CDS \cite{Kim2021}               & 24.61 & 23.77 & 22.68 & 24.51 & 21.96 & 18.95 & 18.27 & 17.60 & 16.46  \\
PCS \cite{Yue2021}               & 34.37 & 28.47 & 23.63 & 17.99 & 16.91 & 15.57 & 17.59 & 16.30 & 15.41 \\
CoDA \cite{Wang2023}             & 18.22 & 17.40 & 16.49 & 19.67 & 17.62 & 15.80 & 15.06 & 14.21 & 13.21 \\
CLIP-UCDIR \cite{Hu2022}         & 24.73 & 21.92 & 20.06 & 28.35 & 27.77 & 26.74 & 35.02 & 33.02 & 30.82 \\
Vanilla CLIP \cite{Radford2021a} & 41.24 & 39.02 & 35.69 & 49.13 & 45.51 & 39.81 & 28.50 & 25.40 & 23.71 \\
CLAIR-A                             &  \textbf{68.92} & \textbf{62.74} & \textbf{55.28} & \textbf{51.53} & \textbf{50.45} & \textbf{48.11} & \textbf{60.34} & \textbf{57.03} & \textbf{52.37} \\
\addlinespace \midrule

\multirow{2}[3]{*}{Method} & \multicolumn{3}{c}{Painting $\rightarrow$ Clipart} & \multicolumn{3}{c}{Clipart $\rightarrow$ Painting} & \multicolumn{3}{c}{Painting $\rightarrow$ Quickdraw} \\
\cmidrule(l){2-10}         & P@50 & P@100 & P@200 & P@50 & P@100 & P@200 & P@50 & P@100 & P@200 \\

\midrule \addlinespace
CDS \cite{Kim2021}               & 37.87 & 35.30 & 32.62 & 39.68 & 35.99 & 33.16 & 8.27 & 8.15 & 8.23 \\
PCS \cite{Yue2021}               & 35.72 & 31.92 & 29.47 & 26.35 & 24.15 & 22.02 & 8.38 & 7.59 & 6.87 \\
CoDA \cite{Wang2023}             & 29.37 & 27.19 & 25.24 & 33.30 & 29.79 & 28.22 & 5.99 & 5.59 & 5.58 \\
CLIP-UCDIR \cite{Hu2022}         & 43.25 & 40.54 & 38.71 & 42.32 & 40.14 & 38.11 & 9.31 & 8.91 & 8.38 \\		
Vanilla CLIP \cite{Radford2021a} & 61.00 & 56.79 & 52.16 & 51.44 & 46.24 & 40.61 & 34.45 & 30.00 & 25.19 \\
CLAIR-A                             & \textbf{74.73} & \textbf{72.95} & \textbf{70.32} & \textbf{81.19} & \textbf{79.49} & \textbf{74.22} & \textbf{44.93} & \textbf{40.38} & \textbf{34.85} \\
\addlinespace \midrule

\multirow{2}[3]{*}{Method} & \multicolumn{3}{c}{Quickdraw $\rightarrow$ Painting} & \multicolumn{3}{c}{Quickdraw $\rightarrow$ Real} & \multicolumn{3}{c}{Real $\rightarrow$ Quickdraw} \\
\cmidrule(l){2-10}         & P@50 & P@100 & P@200 & P@50 & P@100 & P@200 & P@50 & P@100 & P@200 \\

\midrule \addlinespace
CDS \cite{Kim2021}               & 11.28 & 10.38 & 9.27  & 12.60 & 11.40 & 10.02 & 8.14  & 8.07  & 7.99 \\
PCS \cite{Yue2021}               & 6.55  & 6.35  & 6.33  & 13.82 & 12.52 & 11.20 & 11.95 & 10.93 & 9.84 \\
CoDA \cite{Wang2023}             & 7.22  & 6.81  & 6.39  & 7.99  & 7.34  & 6.70  & 6.49  & 6.12  & 5.81 \\
CLIP-UCDIR \cite{Hu2022}         & 10.27 & 9.60  & 9.21  & 10.68 & 9.97  & 9.30  & 9.78  & 9.30  & 8.70 \\
Vanilla CLIP \cite{Radford2021a} & 9.50  & 8.54  & 7.85  & 8.41  & 7.89  & 7.59  & 38.39 & 32.64 & 26.76 \\
CLAIR-A                             & \textbf{26.24} & \textbf{25.61} & \textbf{24.69} & \textbf{26.32} & \textbf{26.47} & \textbf{26.71} & \textbf{51.14}  & \textbf{45.07} & \textbf{37.78}  \\
\addlinespace \midrule

\multirow{6}[6]{*}{Average} & \multicolumn{3}{c}{CDS \cite{Kim2021}} & \multicolumn{3}{c}{PCS \cite{Yue2021}} & \multicolumn{3}{c}{CoDA \cite{Wang2023}} \\
\cmidrule(l){2-10}          & P@50 & P@100 & P@200 & P@50 & P@100 & P@200 & P@50 & P@100 & P@200 \\
\cmidrule(l){2-10}          & 24.93 & 22.83 & 20.69 & 21.30 & 19.29 & 17.54 & 20.18 & 18.23 & 16.62 \\

\cmidrule(l){2-10}          & \multicolumn{3}{c}{CLIP-UCDIR \cite{Hu2022}} & \multicolumn{3}{c}{Vanilla CLIP \cite{Radford2021a}} & \multicolumn{3}{c}{CLAIR-A} \\
\cmidrule(l){2-10}          & P@50 & P@100 & P@200 & P@50 & P@100 & P@200 & P@50 & P@100 & P@200 \\
\cmidrule(l){2-10}          & 26.52 & 24.76 & 23.17 & 42.44 & 38.72 & 34.31 & \textbf{58.33} & \textbf{55.73} & \textbf{51.86} \\

\bottomrule
\end{tabular}

}
\vspace{-1.8mm}
\caption{
Zero-shot Unsupervised Cross-domain Retrieval Accuracy (\%) on test splits of all domain pairings of the DomainNet Zero-Shot dataset. Bold numbers represent best results.
}
\vspace{-1.8mm}
\label{tab:Performance comparison domainnet full}
\end{table*}

\end{document}